\documentclass[runningheads]{llncs}
\usepackage{graphicx}
\usepackage{epsfig}
\usepackage{graphicx}
\usepackage{amsmath}
\usepackage{amssymb}
\usepackage{subfig}
\usepackage{booktabs}

\usepackage{float}
\usepackage{array}
\usepackage{rotating}
\usepackage{makecell}
\usepackage{multirow}

\usepackage{soul}
\usepackage{ulem}

\usepackage{float}

\usepackage{tikz}
\usepackage{comment}
\usepackage{amsmath,amssymb} % define this before the line numbering.
\usepackage{color}
\usepackage[font={small}]{caption}
\usepackage[colorlinks=true,citecolor=green]{hyperref}

\usepackage{amsmath,amsfonts,bm}

\def\eqref#1{equation~\ref{#1}}
\def\1{\bm{1}}

\newcommand{\test}{\mathcal{D_{\mathrm{test}}}}

\DeclareMathAlphabet{\mathsfit}{\encodingdefault}{\sfdefault}{m}{sl}
\SetMathAlphabet{\mathsfit}{bold}{\encodingdefault}{\sfdefault}{bx}{n}

\newcommand{\fulin}{\textcolor{black}}

\newcommand{\daniel}{\textcolor{black}}

\makeatletter
\DeclareRobustCommand\onedot{\futurelet\@let@token\@onedot}
\def\@onedot{\ifx\@let@token.\else.\null\fi\xspace}

\makeatother

\newcommand{\eat}[1]{}
\newcommand{\replace}[2]{} %{\sout{#1} \textbf{#2}}

\begin{document}
% \linenumbers
\pagestyle{headings}
\mainmatter
\def\ECCVSubNumber{014}  % Insert your submission number here

\title{Likelihood Landscapes: A Unifying Principle Behind Many Adversarial Defenses} % Replace with your title

% \setlength{\abovedisplayskip}{3pt}
% \setlength{\belowdisplayskip}{3pt}

\begin{comment}
\titlerunning{ECCV-20 submission ID \ECCVSubNumber} 
\authorrunning{ECCV-20 submission ID \ECCVSubNumber} 
\author{Anonymous ECCV submission}
\institute{Paper ID \ECCVSubNumber}
\end{comment}
%******************

% \begin{comment}
\titlerunning{Likelihood Landscapes}
\author{Fu Lin \and Rohit Mittapalli \\ Prithvijit Chattopadhyay \and Daniel Bolya \and
Judy Hoffman}

\authorrunning{Lin \& Mittapalli et al.}
\institute{Georgia Institute of Technology \\
\email{\{flin68,rmittapalli3,prithvijit3,dbolya,judy\}@gatech.edu}}
% \end{comment}
%******************
\maketitle

% \vspace{\abstractReduceTop}
\begin{abstract}
Convolutional Neural Networks have been shown to be vulnerable to adversarial examples, which are known to locate in subspaces close to where normal data lies but are not naturally occurring and of low probability. In this work, we investigate the potential effect defense techniques have on the geometry of \fulin{the} likelihood landscape - likelihood 
% values over the image space 
of the input images under the trained model. We first propose a way to visualize the likelihood landscape leveraging \fulin{an} energy-based model interpretation of \fulin{discriminative} classifiers. Then we introduce a measure to quantify the flatness of \fulin{the} likelihood landscape. We observe that a subset of adversarial defense techniques results in a similar effect of flattening the likelihood landscape. We further explore directly regularizing \fulin{towards a} flat landscape for adversarial robustness.
% and show that the existing Jacobian regularization work implicitly flattens out the likelihood landscape.
\end{abstract}
% \vspace{\abstractReduceBot}

% \vspace{\sectionReduceTop}
\section{Introduction}
% \vspace{\sectionReduceBot}
Although Convolutional Neural Networks (CNNs) have consistently pushed benchmarks on several computer vision tasks, ranging from 
% image classification \cite{he2016deep}, to object detection \cite{girshick2015fast}, to semantic segmentation \cite{long2015fcn} and to recent multimodal tasks such as visual question answering \cite{antol2015vqa}, visual dialog \cite{visdial} and embodied question answering \cite{embodiedqa}. 
image classification~\cite{he2016deep}, object detection~\cite{girshick2015fast} to recent multimodal tasks such as
visual question answering~\cite{antol2015vqa} and dialog~\cite{visdial},
they are not robust to small adversarial input perturbations.
% Consequentially, this success has enabled deployment of CNNs as critical components of real-world systems such as autonomous vehicles, medical imaging, and retail surveillance. As these systems begin to be deployed beyond virtual domains, it becomes imperative to examine their reliability. 
Prior work has extensively demonstrated the vulnerability of CNNs to adversarial 
attacks~\cite{szegedy2014intriguing,goodfellow2014explaining,papernot2016limitations,carlini2017towards,madry2017towards} and has therefore, exposed how intrinsically unstable these systems are. Countering the susceptibility of CNNs to such attacks has motivated a number of defenses in the computer vision literature~\cite{madry2017towards,zhang2019theoretically,wan2018rethinking,pang2019rethinking,hoffman2019robust,jakubovitz2018improving,samangouei2018defense,raff2019barrage}.

In this work we explore the questions, why are neural networks vulnerable to adversarial attacks in the first place, and how do adversarial defenses protect against them? Are there some inherent 
% exploitable weakness 
deficiencies
in vanilla neural network training that these attacks exploit and that these defenses 
% ``patch up''? 
counter?
To start answering these questions, we explore how adversarial and clean samples fit into the  
% learned data distribution of a model. 
marginal input distributions of trained models.
% In other words, how a model thinks the distribution of input data is.
In doing so, we find that although CNN classifiers implicitly model the distribution of the clean data (both training and test),
% . Standard 
standard training 
% leaves ``holes'' 
induces a specific structure in this 
% \fulin{learned} 
marginal distribution that adversarial attacks can exploit. Moreover, we find that a subset of adversarial defense techniques, despite being wildly different in motivation and implementation, all tend to
% patch these ``holes'', 
modify this structure,
leading to more robust models.

% Explicitly, a 
A standard discriminative CNN-based classifier only explicitly models the 
% distribution of the output classes with respect to the input image 
conditional distribution of the output classes with respect to the input image
(namely, $p_\theta(y | x)$). This is in contrast to generative models (such as GANs \cite{goodfellow2014generative}), which go a step further and model the marginal (or joint) distribution 
% distribution 
of the data directly (namely, $p_\theta(x)$ or $p_{\theta}(x, y)$), so that they can draw new samples from that distribution
or explain existing samples under the learned model. Recently, however, Grathwohl et al. \cite{grathwohl2019your} have shown that it is possible to interpret a CNN classifier as an energy-based model (EBM), 
% without any additional training, 
% directly giving us an formula to compute the likelihood, $p_\theta(x)$. 
allowing us to infer the conditional ($p_\theta(y | x)$) as well as the marginal distribution ($p_\theta(x)$).
% While they use this trick to turn any classifier into a generative model, 
While they use this trick to encode generative capabilities into a discriminative
classifier,
% in this work we use it to explore 
we are interested in exploring 
% the model's implicitly learned marginal data distribution, 
the marginal distribution of the input,
$p_\theta(x)$ for models trained with and without adversarial
defense mechanisms.

\begin{figure}[!t]
    \captionsetup[subfloat]{farskip=2pt,captionskip=0.5pt}
    \begin{center}
    \subfloat[Standard training]{\label{fig:fig1a}\includegraphics[width=0.32\linewidth]{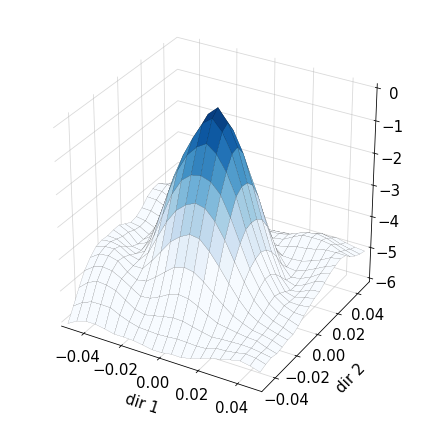}}
    \hfill
    \subfloat[Adv Training, $\epsilon=6$]{\label{fig:fig1b}\includegraphics[width=0.32\linewidth]{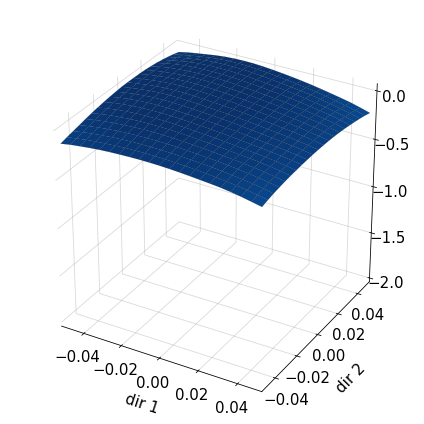}}
    \hfill
    \subfloat[Jacobian Reg, $\lambda=0.05$]{\label{fig:fig1c}\includegraphics[width=0.32\linewidth]{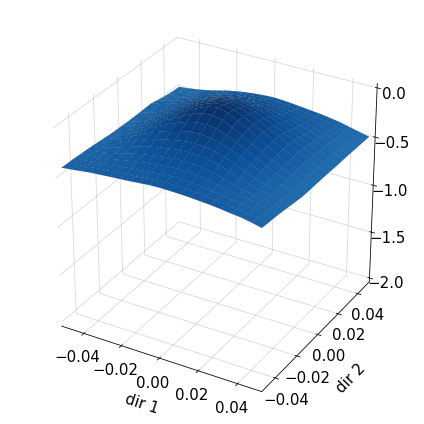}}
    \end{center}
    % \vspace{-15pt}
       \caption{Relative likelihood landscapes $(\Delta \log p_\theta(x))$ plotted over two random perturbation directions for three DDNet models trained on CIFAR10 with the clean sample in the center. (a) follows standard training without additional defense. (b) uses PGD based adversarial training for 10 iterations with $\epsilon=6/255$. (c) uses Jacobian regularization with a strength of $\lambda=0.05$. Models trained to defend against adversarial attacks tend to have much fatter landscapes.}
    \label{fig:fig1}
    % \vspace{-20pt}
\end{figure}

To do this, we 
% devise a novel method to visualize 
study the ``relative likelihood landscape'' ($\Delta \log p_\theta(x)$ in a neighborhood around a test example), which lets us freely analyze this 
% learned distribution. 
marginal distribution.
% And in doing so, 
In doing so, 
% we've noticed 
we notice a worrying trend -- for most training or test examples, a small random perturbation in pixel space can cause the modeled $\log p_\theta(x)$ to drop significantly (see Fig.~\ref{fig:fig1a}). 
% Despite training and test examples (when predicted correctly) having high values for $\log p_\theta(x)$, deviating slightly from these examples knocks the sample out of the models intrinsically learned input distribution.
Despite training and test samples (predicted correctly by the model) being highly likely under the marginal data distribution of the model (high $\log p_\theta(x)$), slight
deviations from these examples significantly moves the perturbed samples
out of the high-likelihood region.
Whether this is because of dataset biases such as chromatic aberration~\cite{doersch2015unsupervised} and JPEG artifacts~\cite{svoboda2016compression} or some other factor~\cite{tramer2019adversarial,engstrom2019adversarial,ilyas2019adversarial}
% (e.g., \daniel{[cite the ``adversarial examples are a property of the data'' paper]} 
isn't clear. Yet, what is clear is that adversarial examples exploit this property (see Fig.~\ref{fig:fig3a}). Moreover, we observe that adversarial defenses intrinsically address this issue.

% Specifically, we find that a subset of adversarial defense techniques --
Specifically, we study two key adversarial defense techniques --
namely adversarial training~\cite{goodfellow2014explaining,madry2017towards,zhang2019theoretically} and Jacobian regularization~\cite{jakubovitz2018improving,hoffman2019robust}. 
% and generative defense approaches~\cite{samangouei2018defense,song2018pixeldefend} in fact share a common pattern. 
Although these 
% adopted 
defense mechanisms are motivated by different objectives, both of them result in elevating the value of $\log p_\theta(x)$ in the region surrounding training and test examples. Thus, they all intrinsically tend to ``flatten'' the likelihood landscape, thereby patching the adversarially-exploitable structures in the network's marginal data distribution (see Fig.~\ref{fig:fig1}). Moreover, we find that the stronger the adversarial defense, the more pronounced this effect becomes (see Fig.~\ref{fig:fig2}).

% To quantify this perceived ``flatness'' of the relative likelihood landscapes, we devise a new metric, $\Phi$-flatness, that captures how close the likelihood of perturbed samples are to the unperturbed ones on average.
To quantify this perceived ``flatness'' of the likelihood
landscapes, we build on top of~\cite{grathwohl2019your} and devise a new metric, $\Phi$-flatness, that captures how rapidly the marginal likelihood
of clean samples change in their immediate neighborhoods.
% And as 
As predicted by our qualitative observations, we find that
stronger adversarial defenses correlate well with higher $\Phi$-flatness and a flatter likelihood landscape (see Fig.~\ref{fig:fig4}). This supports the idea that deviations in the $\log p_\theta(x)$ landscape are connected to what give rise to adversarial examples.

% \sout{\fulin{We also consider directly optimizing for flat landscapes of $\log p_\theta(x)$}}
In order to fully test that hypothesis, we make an attempt to regularize $\log p_\theta(x)$ directly, thereby explicitly enforcing this ``flatness''. 
% After some derivation, we find that surprisingly, 
Our derivations indicate that
regularizing $\log p_\theta(x)$ directly under this model is very similar to Jacobian regularization. We call the resulting regularization scheme AMSReg (based on the Approximate Mass Score proposed in~\cite{grathwohl2019your}) and find that it results in adversarial performance similar to Jacobian regularization. While this defense ends up being less robust than adversarial training, we show that this may simply be because adversarial training prioritizes smoothing out the likelihood landscape in the directions chosen by adversarial attacks (see Fig.~\ref{fig:fig5b}) as opposed to random chosen perturbations. The regularization methods, on the other hand, tend to smooth out the likelihood in random directions very well, but are not as successful in the adversarial directions (see Fig.~\ref{fig:fig5}).
Concretely, we make the following contributions:
\begin{itemize}
\item We 
% design 
propose a way to visualize 
the relative marginal likelihoods of input samples (clean as well as perturbed)
under the trained discriminative
classifier
% a CNN-based classifier's implicitly learned model of the data's distribution, $p_\theta(x)$, 
by leveraging an interpretation of a CNN-based classifier as an energy-based model.
\item We show that the 
% learned 
marginal likelihood of a sample drops significantly with small pixel-level perturbations (Fig.~\ref{fig:fig1a}), and that adversarial examples leverage this property to attack the model (Fig.~\ref{fig:fig3a}). 
\item We empirically identify that a subset of standard defense techniques, including adversarial training and Jacobian regularization, implicitly work to ``flatten'' this 
% learned 
marginal distribution, which addresses this issue (Fig.~\ref{fig:fig1}). The stronger the defense, 
% the 
% more ``flat'' 
``flatter'' the likelihood landscape
% becomes 
(Fig.~\ref{fig:fig2}). 
\item We devise a method for regularizing the ``flatness'' of this likelihood landscape directly (called AMSReg) and arrive at a defense similar to Jacobian regularization that works on par with existing defense methods.
\end{itemize}

% \vspace{\sectionReduceTop}
% \section{Background and Related Work}
\section{Related Work}
% \vspace{\sectionReduceBot}

% \subsection{Adversarial Examples and Attacks}
\par \noindent
\textbf{Adversarial Examples and Attacks.} Adversarial examples for CNNs were originally introduced in Szegedy et al.~\cite{szegedy2014intriguing}, which shows that deep neural networks can be fooled by carefully-crafted imperceptible perturbations. These adversarial examples locate in a subspace that is very close to the naturally occuring data, but they have low probability in the original data distribution~\cite{ma2018characterizing}. Many works have then been proposed to explore ways finding adversarial examples and attacking CNNs. Common attacks include FGSM \cite{goodfellow2014explaining}, JSMA \cite{papernot2016limitations}, C\&W \cite{carlini2017towards}, and PGD \cite{madry2017towards} which exploit the gradient in respect to the input to deceitfully perturb images.

% \subsection{Defenses}
\par \noindent
\textbf{Adversarial Defenses.} Several techniques have also been proposed to defend against adversarial attacks. Some aim at defending models from attack at inference time~\cite{gu2014towards,xie2017mitigating,guo2017countering,samangouei2018defense,raff2019barrage,pang2019mixup}, some aim at detecting adversarial data input~\cite{feinman2017detecting,pang2018towards,lee2018simple,zheng2018robust,qin2019detecting}, and others focus on directly training a model that is robust to a perturbed input~\cite{madry2017towards,zhang2019theoretically,wan2018rethinking,pang2019rethinking,hoffman2019robust,jakubovitz2018improving}.

% Generative approaches are a subset of known inference times defenses. Defense-GAN~\cite{samangouei2018defense} uses a generative adversarial network (GAN) to model the distribution of unperturbed images during training. At inference time, the input images are projected into the learned distribution prior to feeding the images into the classifier. Similarly Pixeldefend~\cite{song2018pixeldefend} approximates the training distribution using a PixelCNN model and purifies perturbed images at inference time by adding small changes to return the images towards a high probability region in the training distribution.

Among those that defend at training time, adversarial training~\cite{goodfellow2014explaining,kurakin2016adversarial,madry2017towards} has been the most prevalent direction. The original implementation of adversarial training~\cite{goodfellow2014explaining} generated adversarial examples using an FGSM attack and incorporated those perturbed images into the training samples. This defense was later enhanced by  Madry et  al.~\cite{madry2017towards} using the stronger projected gradient descent (PGD) attack. However, running strong PGD adversarial training is computationally expensive and several recent works have focused on accelerating the process by reducing the number of attack iterations~\cite{shafahi2019adversarial,zhang2019you,wong2020fast}.

Gradient regularization, on the other hand, adds an additional loss in attempt to produce a more robust model~\cite{drucker1992improving,sokolic2017robust,varga2017gradient,ross2018improving,jakubovitz2018improving,hoffman2019robust}. Of these methods, we consider Jacobian regularization, which tries to minimize the change in output with respect to the change in input (i.e., the Jacobian). In Varga et  al.~\cite{varga2017gradient}, the authors introduce an efficient algorithm to approximate the Frobenius norm of the Jacobian. Hoffman et  al.~\cite{hoffman2019robust} later conducts comprehensive analysis on the efficient algorithm and promotes Jacobian Regularization as a generic scheme for increasing robustness.

\par \noindent
\textbf{Interpreting robustness via geometry of loss landscapes.} In a similar vein to our work, several works~\cite{moosavi2019robustness,qin2019adversarial,yu2019interpreting} have looked at the relation between adversarial robustness and the geometry of \textit{loss landscapes}. Moosavi-Dezfooli et  al.~\cite{moosavi2019robustness} shows that small curvature of loss landscape has strong relation to large adversarial robustness. Yu et  al.~\cite{yu2019interpreting} qualitatively interpret neural network models’ adversarial robustness through loss surface
visualization. In this work, instead of looking at \textit{loss landscapes}, we investigate the relation between adversarial robustness and the geometry of \textit{likelihood landscapes}. The likelihood landscape usually can't be computed for discrimitive models, but given recent work into interpreting neural network classifiers as energy-based models (EBMs) \cite{grathwohl2019your}, we can now study the likelihood of data samples under trained classifiers.

% \vspace{\sectionReduceTop}
% \section{Likelihood Landscape}
\section{Approach and Experiments}
% \vspace{\sectionReduceBot}
We begin our analysis of adversarial examples by exploring the marginal log-likelihood distribution of a clean sample relative to perturbed samples in a surrounding neighborhood (namely, the ``relative likelihood landscape''). Next, we describe how to quantify the ``flatness'' of the landscape and then describe a regularization term in order to optimize for this ``flatness''. Finally, we present our experiment results and observations.

% In this section, we begin by describing how to compute the
% relative likelihood landscape in a neighborhood surrounding the clean sample.
% In what follows, we describe how we quantify ``flatness'' and regularization
% term designed to optimize for ``flatness''. Finally, we present our
% experimental results and observations.
% \vspace{\sectionReduceTop}
% \subsection{How to calculate the log-likelihood $\log p(x)$?}
\subsection{Computing the Marginal Log-Likelihood ($\log p_\theta(x)$)}
% \vspace{\sectionReduceBot}
\label{sec:ebm-related-work}
% In this work, we aim to explore the geometry of landscape of the log-likelihood $\log p(x)$. In the case of generative models, this form of likelihood is readily available. However, \fulin{frequently CNNs are trained discriminatively and hence} can only produce $p_\theta(y|x)$. To get around this limitation (and to compute $p_\theta(x)$ for standard classifiers), we leverage the energy-based model interpretation in Grathwohl et al.~\cite{grathwohl2019your} that gives an expression for $p_\theta(x)$ for any standard classifier.
As mentioned before, CNN-based classifiers are traditionally trained to
model the conditional, 
% likelihood 
% distribution
$\log p_\theta(y|x)$, and not the marginal
likelihood 
% distribution
of samples under the model, $\log p_{\theta}(x)$ (where $\theta$ are the parameters of the model). To get around this restriction, we
leverage the interpretation presented by Grathwohl et al.~\cite{grathwohl2019your}, which
allows us to compute the marginal likelihood based on an energy-based
graphical model interpretation of discriminative classifiers. More
specifically, let $f_\theta:\mathbb{R}^D \rightarrow \mathbb{R}^K$ denote 
% the function within a standard 
a classifier that maps input images to output pre-softmax logits.
% , the core idea is that the logits obtained from $f_\theta$ can be slightly re-interpreted to define $p(x, y)$ and $p(x)$. 
% Concretely, the logits can be used to define an energy based model of the joint distribution of data point $x$ and labels $y$:
% Concretely, 
Grathwohl et al. model the joint
distribution of the input images $x$ and output labels $y$ as:
\begin{align}
    p_\theta(x, y) = \frac{\exp(-E_{\theta}(x,y))}{Z(\theta)} =\frac{\exp({f_{\theta,y}(x)})}{Z(\theta)}~,
    \label{eq:joint}
\end{align}
where $E_{\theta}(x,y)$ is known as the energy function, $f_{\theta,y}(x)$ denotes the $y^{th}$ index of the logits, $f_{\theta}(x)$, and $Z(\theta)=\int_{x}\int_{y}\exp(-E_{\theta}(x,y))$ is an unknown normalizing constant (that depends only on the parameters, and not the input). 
% We can then obtain an unnormalized log-likelihood for $x$ by marginalizing out $y$.
The marginal likelihood of an input sample $x$ can therefore be obtained as,
\begin{align}
    \log p_\theta(x) = \log \sum_y p_\theta(x, y) = \log \left(\sum_y \exp{(f_{\theta, y}(x)})\right) - \log Z(\theta)~.
    \label{eq:logpx}
\end{align}
However, since $Z(\theta)$ involves integrating over the space of all images, it is intractable to compute in practice. Thus, we focus on \textit{relative} likelihoods instead, where we can cancel out the $\log Z(\theta)$ term. Specifically, given a perturbed sample $x'$ and clean sample $x$, we define the relative likelihood of $x'$ w.r.t. $x$ as
\begin{equation}
    \Delta \log p_\theta(x') = \log p_\theta(x') - \log p_\theta(x) = \sum_y \exp{(f_{\theta, y}(x')}) - \sum_y \exp{(f_{\theta, y}(x)})
    \label{eq:rellogpx}
\end{equation}

% \vspace{\sectionReduceTop}
\subsection{Relative Likelihood Landscape Visualization}
% \vspace{\sectionReduceBot}

\begin{figure}[!t]
\captionsetup[subfloat]{farskip=2pt,captionskip=0.5pt}
\begin{center}

\subfloat[Adv Training, $\epsilon=2$]{\label{fig:fig4c}\includegraphics[width=0.32\linewidth]{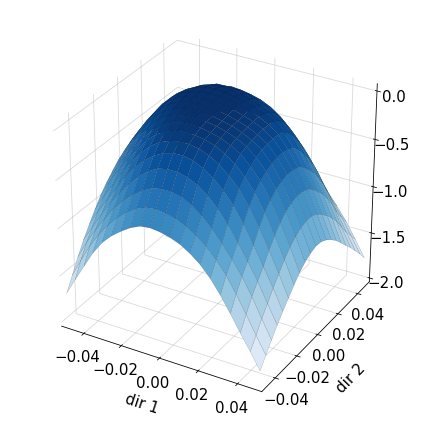}}
\hfill
\subfloat[Adv Training, $\epsilon=4$]{\label{fig:fig4d}\includegraphics[width=0.32\linewidth]{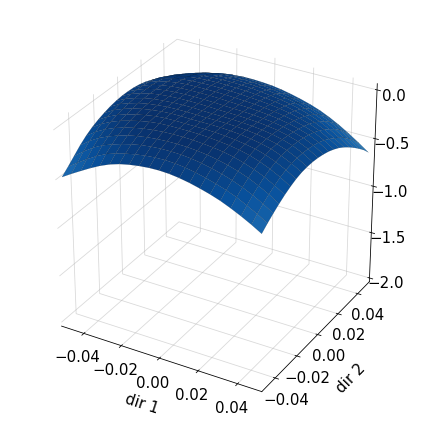}}
\hfill
\subfloat[Adv Training, $\epsilon=6$]{\label{fig:fig4d}\includegraphics[width=0.32\linewidth]{figures/f_cifar_loglandscape_adv6.png}}

\subfloat[Jacobian Reg, $\lambda=0.01$]{\label{fig:fig4c}\includegraphics[width=0.32\linewidth]{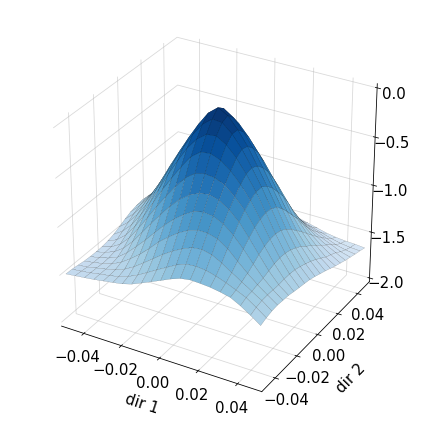}}
\hfill
\subfloat[Jacobian Reg, $\lambda=0.05$]{\label{fig:fig4d}\includegraphics[width=0.32\linewidth]{figures/f_cifar_loglandscape_jaco005.png}}
\hfill
\subfloat[Jacobian Reg, $\lambda=0.1$]{\label{fig:fig4d}\includegraphics[width=0.32\linewidth]{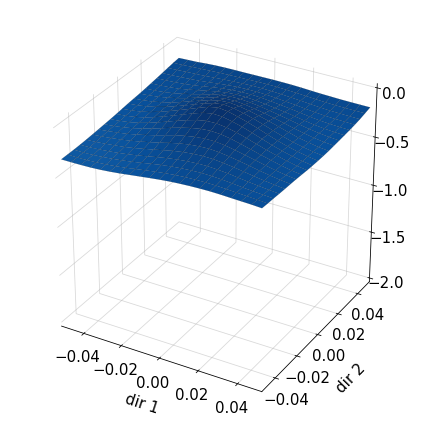}}

\end{center}
% \vspace{-15pt}
   \caption{Relative likelihood landscapes for Adversarial Training and Jacobian regularization with increasing strength. All models used DDNet trained on CIFAR10 and adversarial training uses PGD for 10 iterations. We see that stronger defenses tend to have have flatter landscapes.}
\label{fig:fig2}
% \vspace{-15pt}
\end{figure}
% \prithvi{Building on top of the EBM interpretation of a standard discriminative classifier (see Sec.~\ref{sec:ebm-related-work}) and concepts from prior work focusing on visualizing \textit{loss-landscapes} in either
% the parameter~\cite{goodfellow2014qualitatively,li2018visualizing} or input spaces~\cite{yu2019interpreting},
% we present an approach to visualize the \textit{likelihood landscape} of a trained model.}

We now describe how to utilize Eqn.~\ref{eq:rellogpx} to visualize relative likelihood landscapes, building on top of prior work focusing on visualizing \textit{loss-landscapes} \cite{goodfellow2014qualitatively,li2018visualizing,yu2019interpreting}. To make these plots, we visualize the relative likelihood with respect to a clean sample in a surrounding neighborhood. Namely, for each point $x'$ in the neighborhood $\mathtt{N}(x)$ surrounding a clean sample $x$, we plot $\Delta \log p_\theta(x')$.

\par \noindent
Consistent with prior work, we define the neighborhood $\mathtt{N}(x)$ around a clean sample as
\begin{equation}
    \mathtt{N}(x) = \{ x' | x' =  x + \epsilon_1 d + \epsilon_2 d^{\perp}\}
    \label{eq:neighborhood}
\end{equation}

where $d$ and $d^{\perp}$ are the two randomly pre-selected orthogonal signed vectors, and $\epsilon_1$ and $\epsilon_2$ represent the perturbation intensity along the axes $d$ and $d^{\perp}$. Visualizing $\Delta \log p(x')$ for samples $x'$ surrounding the clean sample $x$ allows us to understand the degree to which structured perturbations affect the likelihood of the sample under the model.

Fig.~\ref{fig:fig1} shows such visualizations on a typical test-time example for standard training (no defense), adversarial training, and training with Jacobian regularization. Note that while we show just one example in these figures, we observe the same general trends over many samples. More examples of these plots can be found in the Appendix. Then as shown in Fig.~\ref{fig:fig1a}, for standard training small perturbations from the center clean image drops the likelihood of that sample considerably. While likelihood doesn't necessarily correlate one to one with accuracy, it's incredibly worrying that standard training doesn't model data samples that are slightly perturbed from test examples. Furthermore, these images are \textit{test} examples that the model hasn't seen during training. This points to an underlying dataset bias, since the model has high likelihood for exactly the test example, but not for samples a small pixel perturbation away. This supports the recent idea that adversarial examples might be more of a property of datasets rather than the model themselves \cite{ilyas2019adversarial}.

The rest of Fig.~\ref{fig:fig1} shows that adversarial defenses like adversarial training and Jacobian regularization tend to have a much more uniform likelihood distribution, where perturbed points have a similar likelihood to the clean sample (i.e., the likelihood landscape is ``flat''). Moreover, we also observe in Fig.~\ref{fig:fig2} that the stronger the defense (for both adversarial training and Jacobian regularization), the lower the variation in the resulting relative likelihood landscape. Specifically, if we increase the adversarial training attack strength ($\epsilon$) or increase the Jacobian regularization strength ($\lambda$), we observe a ``flatter'' likelihood landscape. We measure this correlation quantitatively in Sec.~3.3.

% \footnote{This pattern is reminiscent of how generative approaches like Defense-GAN~\cite{samangouei2018defense} work. Defense-GAN models the data distribution during training and at inference time projects perturbed low-likelihood samples to their nearby high-likelihood samples in the learned distribution. Although not as explicit, it is likely that this implicitly increases the likelihood of the perturbed data, thereby also ``flattening'' the likelihood landscape.}

% In Fig. \ref{fig:fig2}, we visualize the relative likelihood landscapes of models under different defenses near one random test sample (results for more samples can be found in the supplementary document) using two random directions. The central points represent the clean sample. As shown, both adversarial training and Jacobian regularization have a significant effect on flattening the relative likelihood landscape. Moreover, the stronger the regularization, the flatter the relative likelihood landscape.\footnote{\fulin{This pattern is reminiscent of how generative approaches like Defense-GAN~\cite{samangouei2018defense} work. Defense-GAN models the data distribution during training and at inference time projects perturbed low-likelihood samples to their nearby high-likelihood samples in the learned distribution. Although not as explicit, it is likely that this implictly increases the likelihood of the perturbed data, thereby also ``flattening'' the likelihood landscape.}}

\begin{figure}[!t]
\captionsetup[subfloat]{farskip=2pt,captionskip=0.5pt}
\begin{center}
\subfloat[Standard Training]{\label{fig:fig3a}\includegraphics[width=0.5\linewidth]{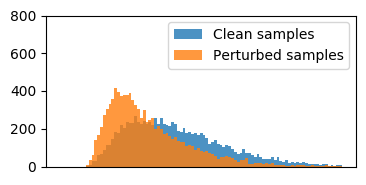}}
\hfill
\subfloat[Jacobian Reg, $\lambda=0.01$]{\label{fig:fig3b}\includegraphics[width=0.5\linewidth]{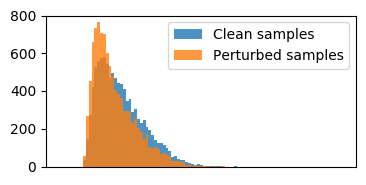}}

\subfloat[Adversarial Training (AT), $\epsilon=2$]{\label{fig:fig3c}\includegraphics[width=0.5\linewidth]{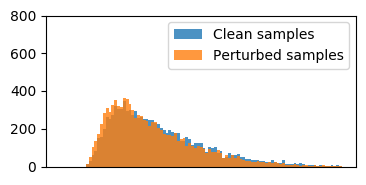}}
\hfill
\subfloat[Adversarial Training (AT), $\epsilon=5$]{\label{fig:fig3d}\includegraphics[width=0.5\linewidth]{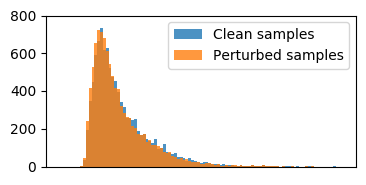}}

\end{center}
% \vspace{-15pt}
   \caption{Histograms of log-likelihood $\log p_\theta(x)$ for clean samples and randomly perturbed samples. Blue corresponds to the log-likelihood on clean samples in CIFAR10 while orange corresponds to the log-likelihood on the perturbed samples. Note that the x-axis for plots might be differently shifted due to the different unknown normalizing constant $Z(\theta)$ for each model, but the scale of x-axis are the same across plots. (a) visualizes DDNet trained without additional defense, (b) uses a Jacobian regularized DDNet and (c)(d) use DDNet trained using PGD based adversarial training for 10 iterations. We see in adversarially robust approaches, clean and perturbed samples have more aligned distributions. We also note that with stronger regularization the distributions become increasingly left-skewed.}
%   \vspace{-15pt}
\label{fig:fig3}
\end{figure}

To get an understanding of the impact different defenses have on the likelihood of overall data distribution, we further plot the histogram of $\log p_\theta(x)$ for all clean test samples and randomly perturbed samples near them (Fig. \ref{fig:fig3}). Adversarial training and Jacobian regularization share several common effects on the log-likelihood distribution: (1) both techniques induce more aligned distribution between clean samples and randomly perturbed samples (likely as a consequence of likelihood landscape flattening). (2) both techniques tend to decrease the log-likelihood of clean samples relatively to randomly perturbed samples. (3) The stronger the regularization, the more left-skewed the log-likelihood distribution of clean samples is.

% Now the question is, is this ``flatness'' necessary for adversarial robustness?

% \vspace{\sectionReduceTop}
% \subsection{Measure for likelihood landscape flatness}
\subsection{Quantifying Flatness}
% \vspace{\sectionReduceBot}
To quantify the degree of variation in the likelihood landscape
surrounding a point, we use the \textit{Approximate Mass Score} (AMS), which was originally introduced in~\cite{grathwohl2019your} as an out of distribution detection score. AMS depends on how the marginal likelihood changes
in the immediate neighborhood of a point in the input space. The Approximate Mass Score can be expressed as,
\begin{align}
    s_\theta(x) = -\left|\left|\frac{\partial \log p_\theta(x)}{\partial x}\right|\right|_F
\end{align}
We use the gradient for all $x' \in \mathtt{N}(x)$ as an indicator of the 
``flatness'' of the likelihood landscape surrounding the clean sample.
% \sout{We design a measure to quantify the flatness of likelihood landscape  of a model based on \textit{Approximate Mass Score}~\cite{grathwohl2019your}, which essentially computes the flatness of likelihood landscape at a specific point of input space,}
% \begin{align}
%     s_\theta(x) = \left|\left|\frac{\partial \log p_\theta(x)}{\partial x}\right|\right|_2
% \end{align}
% Approximate Mass Score (AMS) was originally introduced as a score for out-of-distribution (OOD) detection  \prithvi{but} is \prithvi{related} directly to the flatness of likelihood landscape around $x$. 
% The smaller the Frobenius norm (see Eqn.(?)) is, the flatter the likelihood landscape is in the immediate vicinity of $x$ \prithvi{NOTE -- The proceeding sentence is pointless.}. 
While $\log p_\theta(x)$ is intractable to compute in practice, we can compute $\frac{\partial \log p_\theta(x)}{\partial x}$ exactly,
% {NOTE -- write the expression for the gradient of log likelihood}
% \begin{align}
%     \frac{\partial \log p_\theta(x)}{\partial x}
%     &= \frac{\partial}{\partial x} \log \left(\frac{\sum_y \exp{(f_{\theta, y}(x)})}{Z(\theta)}\right) = \frac{\partial}{\partial x} \log \sum\nolimits_y \exp(f_{\theta, y}(x))
% \end{align}
\begin{align}
    \frac{\partial \log p_\theta(x)}{\partial x}
    &= \frac{\partial}{\partial x} \log \left(\frac{\sum_y \exp{(f_{\theta, y}(x)})}{Z(\theta)}\right) = \sum_y p_{\theta}(y|x) \frac{\partial f_{\theta, y}(x)}{\partial x}
    \label{eq:likelihood_jacobian}
\end{align}
% \sout{One advantage of this measure is that even though the exact value of log-likelihood $\log p_\theta(x)$ is hard to calculate, the exact value of $\frac{\partial \log p_\theta(x)}{\partial x}$ is actually straightforward to compute and also its Forbenius norm.} 
To quantify the flatness in the surrounding area of $x$\daniel{, we sample this quantity at each point in the neighborhood of $x$ (see Eq.~\ref{eq:neighborhood}) and average over $n$ random choices for $d$ and $d^\perp$ to obtain:}
\begin{equation}
    \phi(x) = \sum_{j=1}^{n} \sum_{x' \in \mathtt{N}_j(x)} \frac{s_\theta(x)}{n} = \sum_{j=1}^{n} \sum_{x' \in \mathtt{N}_j(x)} \frac{-1}{n}\left|\left|\frac{\partial \log p_\theta(x')}{\partial x}\right|\right|_F
\end{equation}

\begin{figure}[!t]
\captionsetup[subfloat]{farskip=2pt,captionskip=0.5pt}
\begin{center}
\subfloat[Adversarial training]{\label{fig:fig4a}\includegraphics[width=0.5\linewidth]{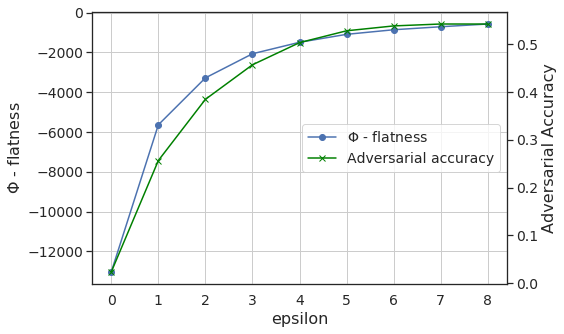}}
\hfill
\subfloat[Jacobian regularization]{\label{fig:fig4b}\includegraphics[width=0.5\linewidth]{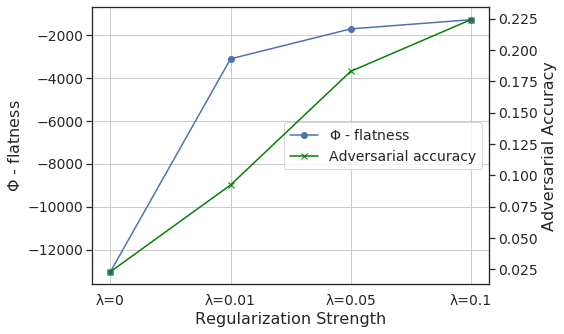}}

\end{center}
% \vspace{-20pt}
   \caption{Visualizes the relationship between the likelihood landscape flatness (as defined by Eqn.~\ref{eq:phi}) and adversarial accuracy. (a) showcases DDNet trained on CIFAR10 using PGD based adversarial training for 10 iterations with increasing epsilon strength. (b) showcases DDNet trained on CIFAR10 using Jacobian regularization of increasing strength. We see that adversarial accuracy and flatness of the relative likelihood landscape are correlated in the case of both defenses.}
\label{fig:fig4}
% \vspace{-15pt}
\end{figure}

The flatness of likelihood landscape for a model is then calculated by averaging $\phi(x)$ for all testing samples.
\begin{equation}
    \Phi = \sum_{i=1}^{N}\frac{\phi(x_i)}{N}
    \label{eq:phi}
\end{equation}
Fig.~\ref{fig:fig4} shows the relationship between flatness score $\Phi$ and defense strength. For both adversarial training (Fig.~\ref{fig:fig4a}) and Jacobian regularization (Fig.~\ref{fig:fig4b}), the stronger the defense, the lower the score and the higher the adversarial accuracy. This observation motivates us to explore directly regularizing this AMS term (AMSReg) to induce a flat likelihood landscape. 

% \vspace{\sectionReduceTop}
\subsection{Approximate Mass Score Regularization (AMSReg)}
% \vspace{\sectionReduceBot}

In our experiments, we observe that high adversarial robustness corresponds with a low Approximate Mass Score in the case of both adversarial training and Jacobian regularization. This begs the question, 
% can we make our models more adversarially 
is it possible to improve adversarial robustness
% robust 
by directly regularizing this term? 
That is, can we regularize $\left|\left|\frac{\partial \log p_\theta(x)}{\partial x}\right|\right|_F^2$ directly in addition to the
cross-entropy loss to encourage robust predictive performance? Note however that naively computing this term 
% explicitly 
during training would require double backpropagation, which would be inefficient. 
% However, if we plug in our assumption of an entropy-based classifier in for $p_\theta(x)$, we find that AMS is essentially a ``weighted'' version of the Jacobian term used in Jacobian regularization. This allows us to regularize the weighted Jacobian term instead (see Prop.~3.1), which can be efficiently implemented with the batch-level Jacobian approximation algorithm proposed in~\cite{hoffman2019robust}.

Instead we make the observation that the right hand side of Eqn.~\ref{eq:likelihood_jacobian} (used to compute $\frac{\partial \log p_\theta(x)}{\partial x}$) is reminiscent of gradient regularization
techniques -- specifically, Jacobian Regularization~\cite{hoffman2019robust}.
% \prithvi{To encourage ``flatness'' in the approximate likelihood landscapes, we also explcitly optimize for AMS. In practice, to encourage "flatness" via a regularizer, instead of AMS, we optimize an upper bound on $\left|\left|\frac{\partial \log p_\theta(x)}{\partial x}\right|\right|_F^2$ (as indicated in Proposition 3.1)
% To encourage flat likelihood landscape, we aim to regularize the Squared Approximate Mass Score $\left|\left|\frac{\partial \log p_\theta(x)}{\partial x}\right|\right|_F^2$. The following proposition provides an upper bound of the squared norm that can be efficiently calculated in batch when using the efficient algorithm proposed in~\cite{hoffman2019robust}.

\par \noindent
\textbf{Connection with Jacobian Regularization~\cite{hoffman2019robust}.} 
% \daniel{Note that Eqn.~\ref{eq:jloss} is very similar to the objective used for Jacobian regularization in~\cite{hoffman2019robust}.
% The difference is that now the Jacobian term is weighted by the predicted probability of each class. Namely, optimizing AMS results in regularizing}
% $||p_\theta (c \mid x) \frac{\partial f_{\theta, c}(x)}{\partial x}||_F^2$, while Jacobian regularization regularizes just $|| \frac{\partial f_{\theta, c}(x)}{\partial x}||_F^2$. Given these expressions, it is not hard to see that the Jacobian regularization term is an upper bound for the AMSReg term. 
As discussed in~\cite{hoffman2019robust}, Jacobian regularization
traditionally emerges as a consequence of stability analysis of model
predictions against input perturbations -- the high-level idea being that
small input perturbations should minimally affect the predictions made
by a network up to a first order Taylor expansion. Hoffman et  al.~\cite{hoffman2019robust}
characterize this by regularizing the Jacobian of the output logits
with respect to the input images. The difference between each element being
summed over in the right hand side of Eqn.~\ref{eq:likelihood_jacobian} and
the input-output Jacobian term used in~\cite{hoffman2019robust} is that now the Jacobian term is weighted by the predicted probability of each class. Namely, optimizing $\left|\left|\frac{\partial \log p_\theta(x)}{\partial x}\right|\right|_F^2$ results in regularizing
$||\sum_c p_\theta (c \mid x) \frac{\partial f_{\theta, c}(x)}{\partial x}||_F^2$, while Jacobian regularization regularizes just $|| \frac{\partial f_{\theta, c}(x)}{\partial x}||_F^2$. The following proposition provides an upper bound of the squared norm that can be efficiently calculated in batch when using the efficient algorithm proposed for Jacobian regularization in~\cite{hoffman2019robust}.
% Therefore, Jacobian regularization can also be seen as implicitly encouraging flat likelihood landscape.

\par \noindent
\textbf{Proposition 3.1.} \textit{Let $C$ denote the number of classes, and $J^w(x) = p_\theta (c \mid x) \frac{\partial f_{\theta, c}(x)}{\partial x_i}$ be a weighted variant of the Jacobian of the class logits with respect to the input. Then,
the frobenius norm of $J^w(x)$ multiplied by the total number of classes upper bounds the
AMS score:}
% \footnote{Please refer to the supplementary document for proof.}
\begin{equation}
    \left|\left|\frac{\partial \log p_\theta(x)}{\partial x}\right|\right|_F^2 \leq C \left|\left|J^w(x)\right|\right|_F^2
\end{equation}

%  PROOF

\par \noindent
\textbf{Proof.} Recall that as per the EBM~\cite{grathwohl2019your}
interpretation of a discriminative classifier, $\frac{\partial \log p_\theta(x)}{\partial x}$ can be expressed as,
% Instead of thinking directly about $\left|\left|\frac{\partial \log p_\theta(x)}{\partial x}\right|\right|_2$, we first think about just $\frac{\partial \log p_\theta(x)}{\partial x}$.
\begin{align}
    \frac{\partial \log p_\theta(x)}{\partial x}
    &= \frac{\partial}{\partial x} \log \left(\frac{\sum_y \exp{(f_{\theta, y}(x)})}{Z(\theta)}\right)\\
    &= \frac{\partial}{\partial x} \log \sum\nolimits_y \exp(f_{\theta, y}(x))\\
    &= \sum\nolimits_y \frac{\exp(f_{\theta, y}(x))}{\sum\nolimits_y \exp(f_{\theta, y}(x))} \frac{\partial f_{\theta,y}(x)}{\partial x}\\
\end{align}
% Therefore, we get
Or,
\begin{equation}
    \frac{\partial \log p_\theta(x)}{\partial x} = \sum\nolimits_y p_\theta(y \mid x) \frac{\partial f_{\theta,y}(x)}{\partial x}
\end{equation}

We utilize the Cauchy-Schwarz inequality, as stated below to obtain an upper bound on $\frac{\partial \log p_\theta(x)}{\partial x}$
\begin{equation}
    (\sum_{k=1}^N u_k v_k)^2 \leq (\sum_{k=1}^N u_k^2) (\sum_{k=1}^N v_k^2)
\end{equation}
where $u_k , v_k \in \mathbb{R} \text{ for all } k \in \{ 1, 2, ..., N \}$. If we set $u_k = 1$ for all $k$, the inequality reduces to,
\begin{equation}
    (\sum_{k=1}^N v_k)^2 \leq N (\sum_{k=1}^N v_k^2)
    \label{eq:cauchy_schwarz_reduction}
\end{equation}
% Using Eqn.~\ref{eq:cauchy_schwarz_reduction}, we can write the following:
Since, $J^w_{y,i}(x) \in \mathbb{R}$, we have,
\begin{equation}
    (\sum_{y=1}^C  J^w_{y,i}(x))^2 \leq C \sum_{y=1}^C  (J^w_{y,i}(x))^2
    \label{eq:cauchy_schwarz_reduction}
\end{equation}
where $C$ is the total number of output classes. This implies that, 
\begin{equation}
    \sum_i (\sum_{y=1}^C  J^w_{y,i}(x))^2 \leq C \sum_i \sum_{y=1}^C  (J^w_{y,i}(x))^2
\end{equation}
Or,
\begin{equation}
    \left|\left|\frac{\partial \log p_\theta(x)}{\partial x}\right|\right|_F^2 \leq C \left|\left|J^w(x)\right|\right|_F^2
\end{equation}

%  PROOF

% where $C$ denotes the number of classes and $J^w(x)$ is a weighted variant of the Jacobian of the logits with respect to the input,
% \begin{equation}
%     J^w_{c,i}(x) = p_\theta (c \mid x) \frac{\partial f_{\theta, c}(x)}{\partial x_i} 
% \end{equation}
% See Appendix for the proof.
% \prithvi{Therefore, the overall objective that we optimize for a mini-batch $B = \{(x,y)\}_{i=1}^{|B|}$ can be expressed as,}
To summarize, the overall objective we optimize for a mini-batch $B = \{(x,y)\}_{i=1}^{|B|}$ using AMSReg can be expressed as, 
\begin{equation}
    \mathcal{L}_{\text{Joint}}(B; \theta) = \sum_{(x,y)\in B} \mathcal{L}_{\text{CE}}(x, y; \theta) + \frac{\mu}{2}\left[\frac{1}{|B|}\sum_{x \in B}\left|\left|J^{w}(x) \right|\right|_F^2\right]
    \label{eq:jloss}
\end{equation}
% We then add $\left|\left|J^w(x)\right|\right|_F^2$ as a loss term in addition to the standard cross-entropy loss. The formula of full loss for a mini-batch $B$ is
% \begin{equation}
%     L_{joint}(\theta) = l(x;y,\theta) + \frac{\mu}{2}[\frac{1}{|B|}\sum_{x \in B}\left|\left|J^{w}(x) \right|\right|_F^2]
%     \label{eq:jloss}
% \end{equation}
where $\mathcal{L}_{\text{CE}}(x, y; \theta)$ is the standard cross-entropy
loss function and $\mu$ is a hyperparameter that determines the strength of regularization.

% Efficient Algorithm Details
\par \noindent
\textbf{Efficient Algorithm for AMSReg.}
Hoffman et al.,~\cite{hoffman2019robust} introduce an efficient algorithm to calculate the Frobenius norm of the input-output Jacobian ($J_{y,i}(x) = \frac{\partial f_{\theta, y}(x)}{\partial x_i}$) using random projection as,
\begin{equation}
    ||J{(x)}||_F^2 = C \mathbb{E}_{\hat{v} \sim S^{C-1}}[||\hat{v} \cdot J||^2]
\end{equation}
where $C$ is the total number of output classes and $\hat{v}$ is random vector drawn from the $(C-1)$-dimensional unit sphere $S^{C-1}$. $||J{(x)}||_F^2$ can then be efficiently approximated by sampling random vectors as,
\begin{equation}
    ||J{(x)}||_F^2 \approx \frac{1}{n_{proj}}\sum^{n_{proj}}_{\mu=1}[\frac{\partial (\hat{v}^{\mu} \cdot z)}{\partial x}]^{2}
\end{equation}
where $n_{proj}$ is the number of random vectors drawn and $z$ is the output vector (see Section 2.3~\cite{hoffman2019robust} for more details).
In AMSReg, we optimize the square of the Frobenius norm of $J^w(x)$ where:
\begin{equation}
    J^w_{y,i}(x) = p_\theta (y \mid x) \frac{\partial f_{\theta, y}(x)}{\partial x_i} 
\end{equation}
Since $J^w{(x)}$ can essentially be expressed as a matrix product of a diagonal matrix of predicted class probabilities and the regular Jacobian matrix, we are able to re-use the efficient random projection algorithm by expressing $||J^w{(x)}||_F^2$ as, 
% which enables us to reuse the efficient random projection algorithm by 
\begin{equation}
    ||J^w{(x)}||_F^2 = C\mathbb{E}_{\hat{v} \sim S^{C-1}}[||\hat{v}^T \mathtt{diag}(p_\theta (c \mid x)) J(x)||^2]
\end{equation}
% Efficient Algorithm Details

% \par \noindent
% \vspace*{-10mm}
% \vspace{\sectionReduceTop}
\subsection{Experiments with AMSReg}
% \vspace{\sectionReduceBot}
% \textbf{Experiment Setup}
We run our experiments on the CIFAR-10 and Fashion-MNIST datasets~\cite{xiao2017fashion}. We
compare AMSReg against both Adversarial Training (AT)~\cite{madry2017towards} and Jacobian Regularization (Jacobian Reg.)~\cite{hoffman2019robust}. We perform our experiments across three network architectures -- LeNet~\cite{LeCun_1998}, DDNet~\cite{papernot2016distillation} and ResNet18~\cite{he2016deep}. We train each model for 200 epochs using piecewise learning rate decay (i.e. decay ten-fold after the 100th and 150th epoch). All models are optimized using SGD with a momentum factor of 0.9. Adversarial training models are trained with 10 steps of PGD. When evaluating adversarial robustness, we use an $l_{\infty}$ bounded 5-steps PGD attack with $\epsilon=8/255$ and a step size of $2/255$ for models trained on CIFAR10 and a 10-step PGD attack with $\epsilon=25/255$ and a step size of $6.25/255$ for models trained on Fashion-MNIST.

\begin{table}[t]
\renewcommand*{\arraystretch}{1.18}
\setlength{\tabcolsep}{3pt}
\begin{center}
\caption{Experiments on CIFAR10: flatness $\Phi$ and accuracies(\%) on clean and adversarial test samples. PGD attacks are generated with $\epsilon=8/255$, steps=5 and a step size of 2/255. AMSReg is able to achieve the highest $\Phi$-Flatness while adversarial training is the most robust defense.}
\resizebox{\columnwidth}{!}{
\begin{tabular}{l l l  c  c c c}
\toprule
& & \textbf{Defense} & & $\Phi$ (\textbf{Flatness} $\uparrow$) & \textbf{Clean-Acc}. $\uparrow$ (\%) & \textbf{Adversarial-Acc.} $\uparrow$ (\%) (PGD) \\
\midrule
\multirow{4}{*}{\rotatebox{90}{\centering DDNet~\cite{papernot2016distillation} }} &
 & No Defense && $-13035.5$ & $\mathbf{91.45}$ & $1.7$ \\
& & AT ($\epsilon=3/255$) && $-2066.4$ & $85.90$ & $\mathbf{45.71}$ \\
& & Jacobian Reg.~\cite{hoffman2019robust} ($\lambda=0.05$) && $-1692.6$ & $86.06$ & $18.31$ \\
& & AMSReg ($\lambda=1.0$) && $\mathbf{-768.7}$ & $85.45$ & $20.78$ \\
\midrule
\multirow{4}{*}{\rotatebox{90}{\centering RN-18~\cite{he2016deep} }} &
 & No Defense && $-13974.3$ & $\mathbf{94.69}$ & $0.1$ \\
& & AT ($\epsilon=3/255$) && $-1055.0$ & $85.32$ & $\mathbf{57.48}$ \\
& & Jacobian Reg.~\cite{hoffman2019robust} ($\lambda=0.05$) && $-1945.1$ & $82.28$ & $14.73$ \\
& & AMSReg ($\lambda=1.0$) && $\mathbf{-659.5}$ & $86.94$ & $11.07$ \\
\bottomrule
\end{tabular}}
\label{tbl:table1}
\end{center}
\vspace{-20pt}
\end{table}

\par \noindent
\textbf{Results and Analysis.} Table~\ref{tbl:table1} and Table~\ref{tbl:table2} show the results of AMSReg compared to adversarial training and Jacobian regularization on CIFAR10 and Fashion-MNIST. It is known that there is a trade-off between adversarial robustness and accuracy~\cite{zhang2019theoretically,tsipras2018robustness}. Therefore, here we only compare models that have similar accuracy on clean test images. As observed in Table \ref{tbl:table1} and \ref{tbl:table2}, models trained with AMSReg indeed have flatter likelihood landscapes than Jacobian regularization and adversarial training, \daniel{indicating that the regularization was successful.} 
% And as hinted by the math, }the 
Furthermore, AMSReg models have comparable adversarial robustness to that of Jacobian regularization. However, they are still significantly 
% weaker 
susceptible
to adversarial attacks than adversarial training.
% than that of adversarial training.

\begin{table}[t]
\renewcommand*{\arraystretch}{1.18}
\setlength{\tabcolsep}{3pt}
\begin{center}
\caption{Experiments on Fashion-MNIST: flatness $\Phi$ and accuracies(\%) on clean and adversarial test samples. PGD attacks are generated with $\epsilon=25/255$, steps=10 and a step size of 6.25/255. AMSReg is able to achieve the highest $\Phi$-Flatness while adversarial training is the most robust defense.}
\resizebox{\columnwidth}{!}{
\begin{tabular}{l l l  c  c c c}
\toprule
& & \textbf{Defense} & & $\Phi$ (\textbf{Flatness} $\uparrow$) & \textbf{Clean-Acc}. $\uparrow$ (\%) & \textbf{Adversarial-Acc.} $\uparrow$ (\%) (PGD) \\
\midrule
\multirow{4}{*}{\rotatebox{90}{\centering LeNet~\cite{LeCun_1998} }} &
 & No Defense && $-9435.9$  & $\mathbf{91.53}$ & $0.0$ \\
& & AT ($\epsilon=10/255$) && $-1504.0$ & $89.35$ & $\mathbf{62.64}$ \\
& & Jacobian Reg.~\cite{hoffman2019robust} ($\lambda=0.01$) && $-1694.6$ & $88.76$ &$34.71$ \\
& & AMSReg ($\lambda=0.5$) && $\mathbf{-378.1}$ & $89.51$ & $34.04$ \\
\midrule
\multirow{4}{*}{\rotatebox{90}{\centering RN-18~\cite{he2016deep} }} &
 & No Defense && $-12952.8$ & $\mathbf{93.75}$ & $0.1$ \\
& & AT ($\epsilon=10/255$) && $-1257.8$ & $91.64$ & $\mathbf{65.24}$ \\
& & Jacobian Reg.~\cite{hoffman2019robust} ($\lambda=0.01$) && $-821.7$ & $91.04$ & $45.37$ \\
& & AMSReg ($\lambda=0.5$) && $\mathbf{-482.9}$ & $91.29$ & $46.06$ \\
\bottomrule
\end{tabular}}
\label{tbl:table2}
\end{center}
\vspace{-20pt}
\end{table}

\begin{figure}[h!]
\captionsetup[subfloat]{farskip=2pt,captionskip=0.5pt}
\begin{center}

\subfloat[Standard training]{\label{fig:fig5a}\includegraphics[width=0.25\linewidth]{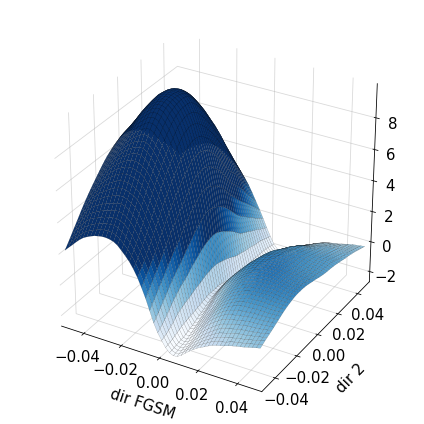}}
\hfill
\subfloat[Adv Train, $\epsilon=6$]{\label{fig:fig5b}\includegraphics[width=0.25\linewidth]{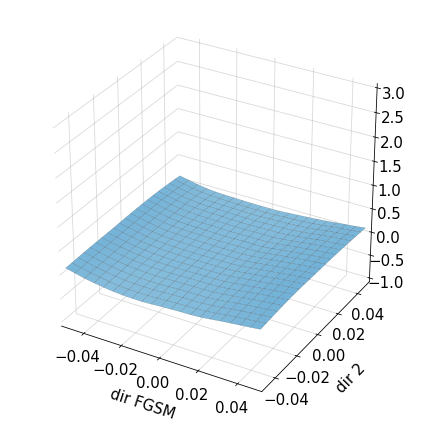}}
\hfill
\subfloat[AMSReg, $\lambda=1$]{\label{fig:fig5c}\includegraphics[width=0.25\linewidth]{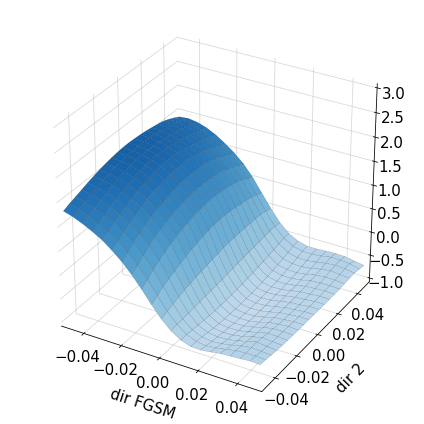}}
\hfill
\subfloat[Jacobian, $\lambda=0.1$]{\label{fig:fig5d}\includegraphics[width=0.25\linewidth]{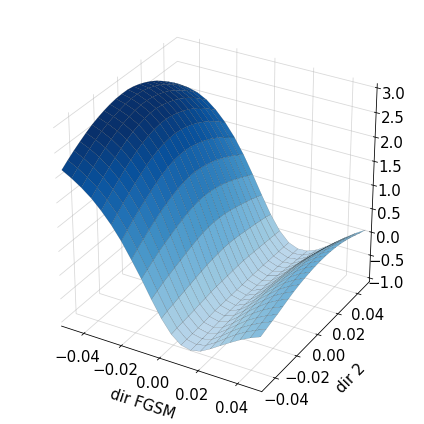}}

\end{center}
% \vspace{-15pt}
   \caption{Relative likelihood landscapes projected using one FGSM direction (\texttt{dir FGSM}) and a random direction (\texttt{dir2}). Each visualization uses DDNet trained on CIFAR10. (a) uses standard training with no defense, (b) is adversarially trained with a 10-step PGD attack, (c) uses AMSReg with $\lambda=1$, and (d) uses Jacobian Regularization with $\lambda = 0.1$. We see that adversarial training is able to flatten the likelihood landscape in the FGSM direction much more than the regularization defenses.}
\label{fig:fig5}
% \vspace{-15pt}
\end{figure}

We conduct further analysis by looking at the flatness along the attack directions (i.e., the direction an FGSM attack chosen to perturb the image) instead of random directions. Fig.~\ref{fig:fig5} shows the likelihood landscape projected using one FGSM direction and one random direction. \daniel{Interestingly enough, both Jacobian and AMS regularization are able to flatten the landscape in the random direction well (with AMS being slightly flatter than Jacobian). However, neither regularization method is able to flatten out the FGSM attack direction nearly as much as adversarial training.}
% It's interesting that even though AMSReg is able to flatten out random directions better than adversarial training, adversarial training tends to flatten out in the FGSM direction significantly better than AMSReg
This might explain why neither AMS nor Jacobian regularization are able to match the adversarial robustness of adversarial training. We leave the next step of flattening out the worst-case directions (i.e. the attack directions) for future work.

% \section{Experiments}
% \input{sections/experiments}

% \vspace{\sectionReduceTop}
\section{Conclusion}
% \vspace{\sectionReduceBot}
In this work, we explore why neural networks are vulnerable to adversarial attacks from a perspective of the marginal distribution of the inputs (images) under the trained model. We first suggest a way to visualize the likelihood landscape of CNNs by leveraging the recently proposed EBM interpretation of a discriminatively learned classifier. Qualitatively, we show that a subset of standard defense techniques such as adversarial training and Jacobian regularization share a common pattern: they induce flat likelihood landscapes. We then quantitatively show this correlation by introducing a measure regarding the flatness of the likelihood landscape in the surrounding area of clean samples. We also explore directly regularizing a term that encourages flat likelihood landscapes, but this results in worse adversarial robustness than adversarial training and is roughly comparable to just Jacobian regularization. After further analysis, we find that adversarial training significantly flattens the likelihood landscape in the directions abused by adversarial attacks. The regularization methods, on the other hand, are better at flattening the landscape in random directions. These findings suggest that flattening the likelihood landscape is important for adversarial robustness, but it's most important for the landscape to be flat in the directions chosen by adversarial attacks.

% After analysis, we find that adversarial training out-perform our regularizer in flattening out attack directions while Jacobian regularization and AMSReg are better at flattening out random directions. 
Overall, our findings in this paper provide a new perspective of adversarial robustness as flattening the likelihood landscape and show how different defenses and regularization techniques address this similar core deficiency. We leave designing a regularizer that flattens the likelihood landscape in attack directions for future investigation.
% Designing a regularizer that can flatten out attack directions is a direction for future investigation.

% \section{Conclusions}

% \clearpage
\bibliographystyle{splncs04}
\bibliography{egbib,main}

\clearpage
\appendix
\section{Appendix}
In this section, we first provide some details on the efficient algorithm used to compute AMSReg (see Section 3.4, main paper) during training. Next, we present some additional examples of our relative likelihood
landscape visualization.
\subsection{Details on the efficient algorithm implementation for AMSReg.}
Hoffman et al.,~\cite{hoffman2019robust} introduce an efficient algorithm to calculate the Frobenius norm of the input-output Jacobian ($J_{y,i}(x) = \frac{\partial f_{\theta, y}(x)}{\partial x_i}$) using random projection as,
\begin{equation}
    ||J{(x)}||_F^2 = C \mathbb{E}_{\hat{v} \sim S^{C-1}}[||\hat{v} \cdot J||^2]
\end{equation}
where $C$ is the total number of output classes and $\hat{v}$ is random vector drawn from the $(C-1)$-dimensional unit sphere $S^{C-1}$. $||J{(x)}||_F^2$ can then be efficiently approximated by sampling random vectors as,
\begin{equation}
    ||J{(x)}||_F^2 \approx \frac{1}{n_{proj}}\sum^{n_{proj}}_{\mu=1}[\frac{\partial (\hat{v}^{\mu} \cdot z)}{\partial x}]^{2}
\end{equation}
where $n_{proj}$ is the number of random vectors drawn and $z$ is the output vector (see Section 2.3~\cite{hoffman2019robust} for more details).
In AMSReg, we optimize the square of the Frobenius norm of $J^w(x)$ where:
\begin{equation}
    J^w_{y,i}(x) = p_\theta (y \mid x) \frac{\partial f_{\theta, y}(x)}{\partial x_i} 
\end{equation}
Since $J^w{(x)}$ can essentially be expressed as a matrix product of a diagonal matrix of predicted class probabilities and the regular Jacobian matrix, we are able to re-use the efficient random projection algorithm by expressing $||J^w{(x)}||_F^2$ as, 
% which enables us to reuse the efficient random projection algorithm by 
\begin{equation}
    ||J^w{(x)}||_F^2 = C\mathbb{E}_{\hat{v} \sim S^{C-1}}[||\hat{v}^T \mathtt{diag}(p_\theta (c \mid x)) J(x)||^2]
\end{equation}

\subsection{More examples of likelihood landscape visualization}

\begin{figure}
\captionsetup[subfloat]{farskip=2pt,captionskip=0.5pt}
\begin{center}

\subfloat[Standard training]{\includegraphics[width=0.32\linewidth]{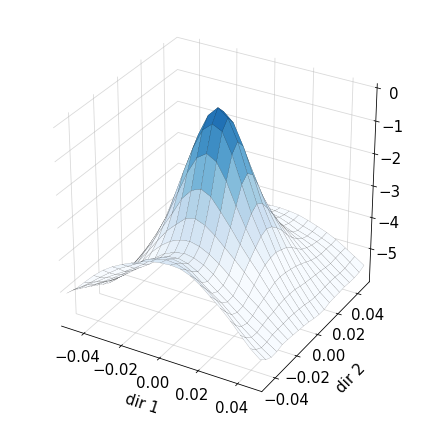}}
\hfill
\subfloat[Jacobian Reg, $\lambda=0.01$]{\includegraphics[width=0.32\linewidth]{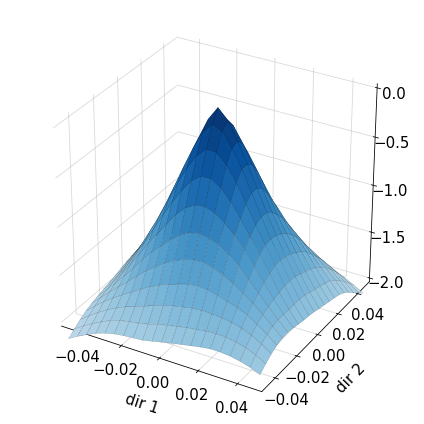}}
\hfill
\subfloat[Jacobian Reg, $\lambda=0.1$]{\includegraphics[width=0.32\linewidth]{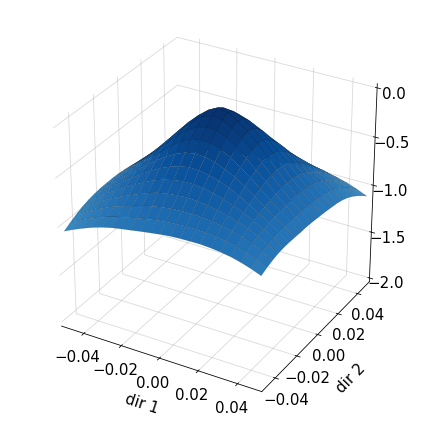}}

\subfloat[Adv Training, $\epsilon=2$]{\includegraphics[width=0.32\linewidth]{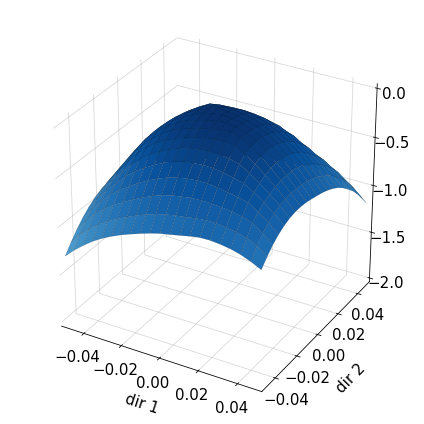}}
\hfill
\subfloat[Adv Training, $\epsilon=4$]{\includegraphics[width=0.32\linewidth]{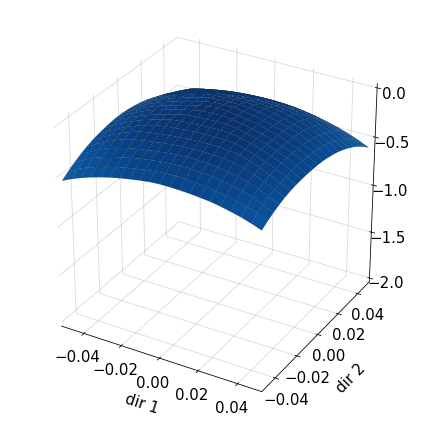}}
\hfill
\subfloat[Adv Training, $\epsilon=8$]{\includegraphics[width=0.32\linewidth]{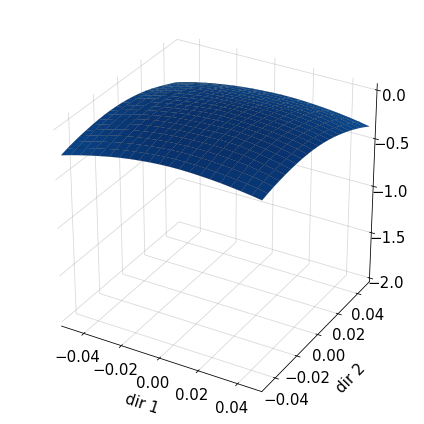}}

\end{center}
   \caption{Relative likelihood landscapes for Adversarial Training and Jacobian regularization with increasing strength. All models used DDNet trained on CIFAR10 and adversarial training uses PGD for 10 iterations.}
\label{fig:fig2_1}
\end{figure}

\begin{figure}
\captionsetup[subfloat]{farskip=2pt,captionskip=0.5pt}
\begin{center}

\subfloat[Standard training]{\includegraphics[width=0.32\linewidth]{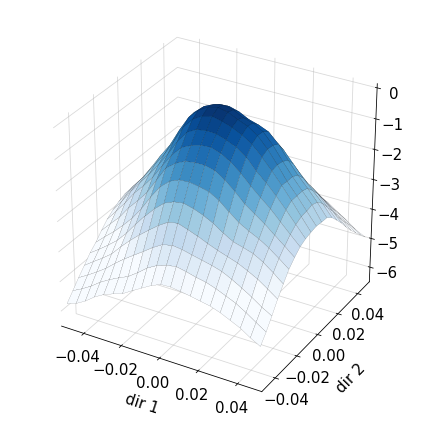}}
\hfill
\subfloat[Jacobian Reg, $\lambda=0.01$]{\includegraphics[width=0.32\linewidth]{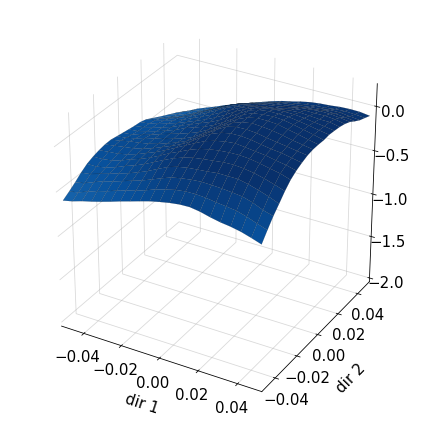}}
\hfill
\subfloat[Jacobian Reg, $\lambda=0.1$]{\includegraphics[width=0.32\linewidth]{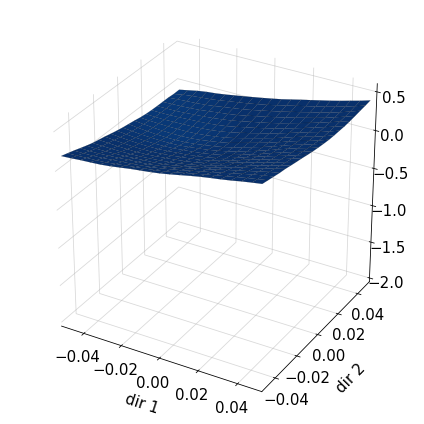}}

\subfloat[Adv Training, $\epsilon=2$]{\includegraphics[width=0.32\linewidth]{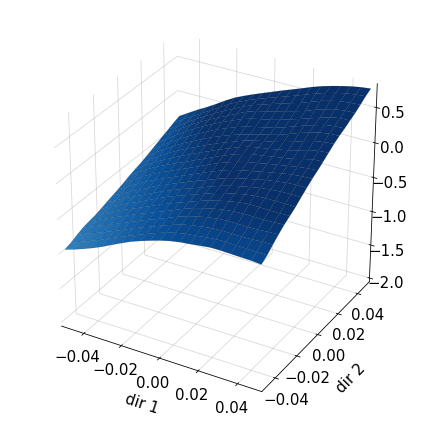}}
\hfill
\subfloat[Adv Training, $\epsilon=4$]{\includegraphics[width=0.32\linewidth]{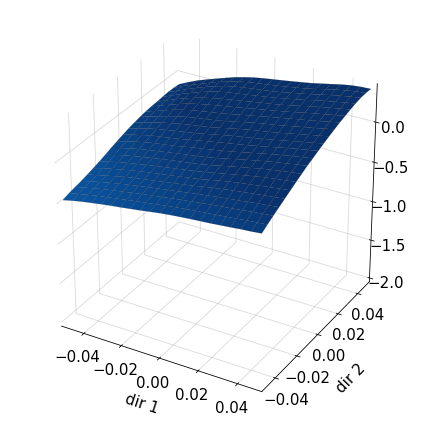}}
\hfill
\subfloat[Adv Training, $\epsilon=8$]{\includegraphics[width=0.32\linewidth]{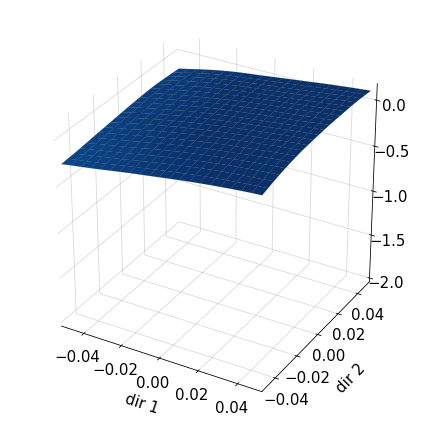}}

\end{center}
   \caption{Relative likelihood landscapes for Adversarial Training and Jacobian regularization with increasing strength. All models used DDNet trained on CIFAR10 and adversarial training uses PGD for 10 iterations.}
\label{fig:fig2_2}
\end{figure}

\begin{figure}
\captionsetup[subfloat]{farskip=2pt,captionskip=0.5pt}
\begin{center}

\subfloat[Standard training]{\includegraphics[width=0.32\linewidth]{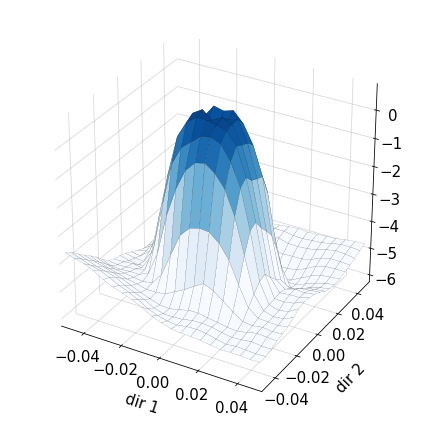}}
\hfill
\subfloat[Adv Training, $\epsilon=1$]{\includegraphics[width=0.32\linewidth]{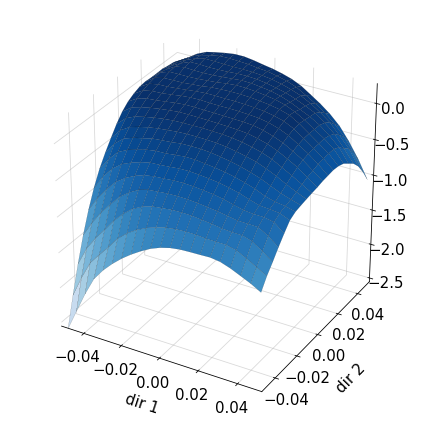}}
\hfill
\subfloat[Adv Training, $\epsilon=3$]{\includegraphics[width=0.32\linewidth]{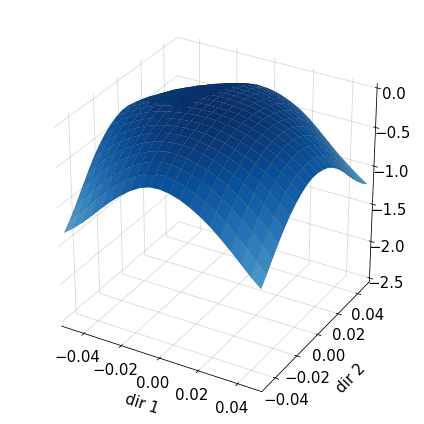}}

\subfloat[Jacobian Reg, $\lambda=0.01$]{\includegraphics[width=0.32\linewidth]{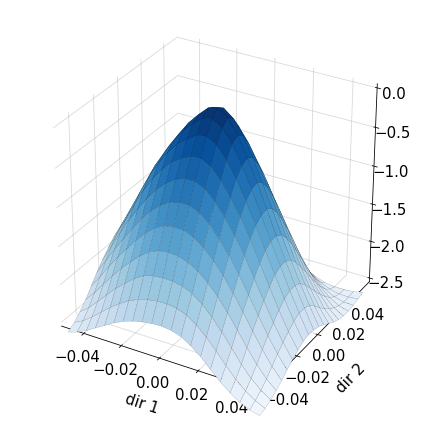}}
\hfill
\subfloat[Jacobian Reg, $\lambda=0.05$]{\includegraphics[width=0.32\linewidth]{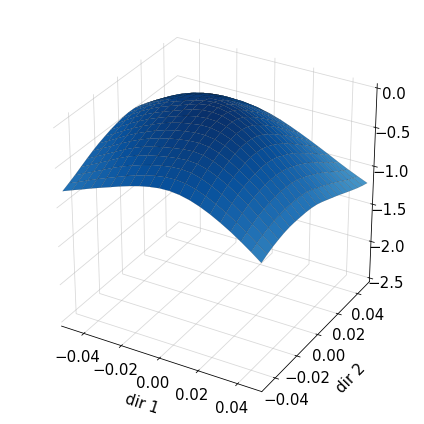}}
\hfill
\subfloat[Jacobian Reg, $\lambda=0.1$]{\includegraphics[width=0.32\linewidth]{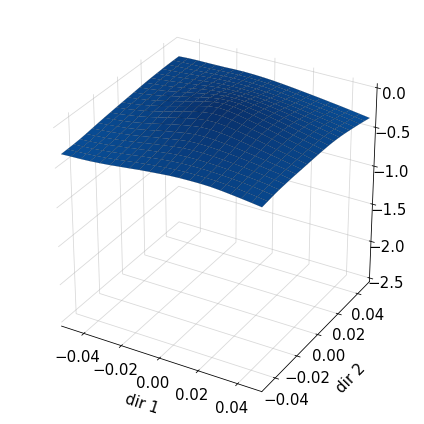}}

\end{center}
   \caption{Relative likelihood landscapes for Adversarial Training and Jacobian regularization with increasing strength. All models used ResNet18 trained on CIFAR10 and adversarial training uses PGD for 10 iterations.}
\label{fig:fig2_3}
\end{figure}

We show more examples of relative likelihood landscape visualization that are similar to Fig. 2 in the main paper but are for different test samples and network architectures. Concretely, Fig. \ref{fig:fig2_1} and Fig. \ref{fig:fig2_2} shows the relative likelihood landscapes for two test samples of CIFAR10 using DDNet~\cite{papernot2016distillation} model trained with different defenses. Fig. \ref{fig:fig2_3} shows the relative likelihood landscapes for one test sample of CIFAR10 using ResNet18~\cite{he2016deep} model trained with different defenses. Overall, we can see that stronger defenses tend to have flatter landscapes.

% \begin{figure}[!H]

\end{document}